\documentclass[10pt,twocolumn,letterpaper]{article}

\usepackage{cvpr}
\usepackage{times}
\usepackage{epsfig}
\usepackage{graphicx}
\usepackage{amsmath}
\usepackage{amssymb}

\usepackage{enumitem}
\usepackage{booktabs}
\usepackage{multirow}
\usepackage{subcaption}
\usepackage{mmstyle}

\usepackage[breaklinks=true,bookmarks=false,colorlinks]{hyperref}

\cvprfinalcopy 

\def\cvprPaperID{****} 

\ifcvprfinal\fi
\begin{document}

\title{Hybrid Task Cascade for Instance Segmentation}

\author{Kai Chen$^1$ \quad Jiangmiao Pang$^{2,3}$ \quad Jiaqi Wang$^1$ \quad Yu Xiong$^1$ \quad Xiaoxiao Li$^1$ \quad Shuyang Sun$^4$\\
Wansen Feng$^2$ \quad Ziwei Liu$^1$ \quad Jianping Shi$^2$ \quad Wanli Ouyang$^4$ \quad Chen Change Loy$^5$ \quad Dahua Lin$^1$\\
$^1$The Chinese University of Hong Kong \quad $^2$SenseTime Research \quad $^3$Zhejiang University \\
$^4$The University of Sydney \quad $^5$Nanyang Technological University\\
}

\maketitle
\thispagestyle{empty}


\begin{abstract}

Cascade is a classic yet powerful architecture that has boosted performance on
various tasks. However, how to introduce cascade to instance segmentation remains
an open question. A simple combination of Cascade R-CNN and Mask R-CNN
only brings limited gain.
In exploring a more effective approach, we find that the key to a successful
instance segmentation cascade is to fully leverage the reciprocal relationship
between detection and segmentation.
In this work, we propose a new framework, Hybrid Task Cascade (HTC), which
differs in two important aspects:
(1) instead of performing cascaded refinement on these two tasks separately,
it interweaves them for a joint multi-stage processing;
(2) it adopts a fully convolutional branch to provide spatial context, which can
help distinguishing hard foreground from cluttered background.
Overall, this framework can learn more discriminative features
progressively while integrating complementary features together in each stage.
Without bells and whistles, a single HTC obtains
38.4\% and 1.5\% improvement over a strong Cascade Mask R-CNN
baseline on MSCOCO dataset.
Moreover, our overall system achieves 48.6 mask AP on the test-challenge split, ranking 1st in the COCO 2018 Challenge Object Detection Task.
Code is available at: \url{https://github.com/open-mmlab/mmdetection}.

\end{abstract}


\section{Introduction}
\label{sec:intro}

Instance segmentation is a fundamental computer vision task that performs
per-pixel labeling of objects at instance level. Achieving accurate and robust
instance segmentation in real-world scenarios such as autonomous driving and
video surveillance is challenging.
Firstly, visual objects are often subject to deformation, occlusion and scale
changes. Secondly, background clutters make object instances hard to be
isolated. To tackle these issues, we need a robust representation that is
resilient to appearance variations. At the same time, it needs to capture rich
contextual information for discriminating objects from cluttered background.

Cascade is a classic yet powerful architecture that has boosted performance on
various tasks by multi-stage refinement. Cascade R-CNN~\cite{cai18cascadercnn}
presented a multi-stage architecture for object detection and achieved promising
results. The success of Cascade R-CNN can be ascribed to two key aspects:
(1) progressive refinement of predictions and
(2) adaptive handling of training distributions.

Though being effective on detection tasks, integrating the idea of cascade into
instance segmentation is nontrivial. A direct combination of Cascade R-CNN and
Mask R-CNN~\cite{he2017mask} only brings limited gain in terms of mask AP
compared to bbox AP. Specifically, it improves bbox AP by $3.5\%$ but mask AP by
$1.2\%$, as shown in Table~\ref{tab:overall-results}. An important reason for
this large gap is the suboptimal information flow among mask branches of
different stages. Mask branches in later stages only benefit from better
localized bounding boxes, without direct connections.

To bridge this gap, we propose Hybrid Task Cascade (HTC),
a new cascade architecture for instance segmentation.
The key idea is to improve the information flow by incorporating cascade and
multi-tasking at each stage and leverage spatial context to further boost the
accuracy.
Specifically, we design a cascaded pipeline for progressive refinement.
At each stage, both bounding box regression and mask prediction
are combined in a multi-tasking manner.
Moreover, \emph{direct} connections are introduced between the mask branches
at different stages -- the mask features of each stage will be embedded and
fed to the next one, as demonstrated in Figure~\ref{fig:mask-arch}.
The overall design strengthens the information flow between tasks and
across stages, leading to better refinement at each stage and more accurate
predictions on all tasks.


For object detection, the scene context also provides useful clues,
\eg~for inferring the categories, scales, etc. To leverage this context, we
incorporate a fully convolutional branch that performs pixel-level stuff
segmentation. This branch encodes contextual information, not only from foreground
instances but also from background regions, thus complementing the bounding boxes
and instance masks. Our study shows that the use of the spatial contexts helps
to learn more discriminative features.

HTC is easy to implement and can be trained end-to-end.
Without bells and whistles, it achieves $2.6\%$ and $1.4\%$ higher mask AP than
Mask R-CNN and Cascade Mask R-CNN baselines respectively on the challenging COCO dataset.
Together with better backbones and other common components, \eg.~deformable convolution,
multi-scale training and testing, model ensembling, we achieve $49.0$ mask AP on test-dev dataset,
which is 2.3\% higher than the winning approach~\cite{liu2018path} of COCO Challenge 2017.

Our main contributions are summarized as follows:
(1) We propose Hybrid Task Cascade (HTC), which effectively integrates cascade into instance segmentation by interweaving detection and segmentation features together for a joint multi-stage processing.
It achieves the state-of-the-art performance on COCO test-dev and test-challenge.
(2) We demonstrate that spatial contexts benefit instance segmentation by discriminating foreground objects from background clutters.
(3) We perform extensive study on various components and designs, which provides a reference and is helpful for futher research on object detection and instance segmentation.



\section{Related Work}

\paragraph{Instance Segmentation.}
Instance segmentation is a task to localize objects of interest in an image
at the pixel-level, where segmented objects are generally represented by
masks. This task is closely related to both object detection and semantic
segmentation~\cite{liu2015semantic, li2017not}. Hence, existing methods for this task roughly fall into
two categories, namely detection-based and segmentation-based.

Detection-based methods resort to a conventional detector to generate bounding
boxes or region proposals, and then predict the object masks within the
bounding boxes.
Many of these methods are based on CNN, including
DeepMask~\cite{pinheiro2015learning},
SharpMask~\cite{pinheiro2016learning},
and InstanceFCN~\cite{dai2016instance}.
MNC~\cite{dai2016mnc} formulates instance segmentation as a pipeline
that consists of three sub-tasks: instance localization, mask
prediction and object categorization, and trains the whole network end-to-end
in a cascaded manner.
In a recent work, FCIS~\cite{li2017fully} extends InstanceFCN and presents a
fully convolutional approach for instance segmentation.
Mask-RCNN~\cite{he2017mask} adds an extra branch based on Faster
R-CNN~\cite{ren2015faster} to obtain pixel-level mask predictions,
which shows that a simple pipeline can yield promising results.
PANet~\cite{liu2018path} adds a bottom-up path besides the top-down path in
FPN~\cite{lin2017feature} to facilitate the information flow.
MaskLab~\cite{chen2018masklab} produces instance-aware masks by combining
semantic and direction predictions.

Segmentation-based methods, on the contrary, first obtains a pixel-level
segmentation map over the image, and then identifies object instances therefrom.
Along this line,
Zhang \etal~\cite{zhang2015monocular, zhang2016instance}
propose to predict instance labels based on local patches and integrate
the local results with an MRF.
Arnab and Torr~\cite{arnab2016bottom} also use CRF to identify instances.
Bai and Urtasun~\cite{bai2017deep} propose an alternative way, which
combines watershed transform and deep learning to produce an energy map,
and then derive the instances by dividing the output of the watershed transform.
Other approaches include bridging category-leval and instance-level
segmentation~\cite{wu2016bridging}, learning a boundary-aware mask
representation~\cite{hayder2017boundary}, and employing a sequence of
neural networks to deal with different sub-grouping problems~\cite{liu2017sgn}.

\paragraph{Multi-stage Object Detection.}
The past several years have seen remarkable progress in object detection.
Mainstream object detection frameworks are often categorized into two types,
single-stage detectors~\cite{liu2016ssd,redmon2016you,lin2017focal}
and two-stage detectors~\cite{ren2015faster,dai16rfcn,he2017mask,Libra2019CVPR}.
Recently, detection frameworks with multiple stages emerge as an increasingly
popular paradigm for object detection.
Multi-region CNN~\cite{gidaris2015object} incorporates an iterative
localization mechanism that alternates between box scoring and location
refinement.
AttractioNet~\cite{gidaris2016attend} introduces an Attend \& Refine module
to update bounding box locations iteratively.
CRAFT~\cite{binyang16craft} incorporates a cascade structure into
RPN~\cite{ren2015faster} and Fast R-CNN~\cite{girshick2015fast} to improve
the quality of the proposal and detection results.
IoU-Net~\cite{jiang2018acquisition} performs progressive bounding box refinement
(even though not presenting a cascade structure explicitly).
Cascade structures are also used to exclude easy negative samples.
For example,
CC-Net~\cite{ouyang2017learning} rejects easy RoIs at shallow layers.
Li \etal~\cite{li2015convolutional} propose to operate at multiple resolutions
to reject simple samples.
Among all the works that use cascade structures,
Cascade R-CNN~\cite{cai18cascadercnn} is perhaps the most relevant to ours.
Cascade R-CNN comprises multiple stages, where the output of each stage is fed
into the next one for higher quality refinement.
Moreover, the training data of each stage is sampled with increasing IoU
thresholds, which inherently handles different training distributions.

While the proposed framework also adopts a cascade structure, it differs
in several important aspects. First, multiple tasks, including detection,
mask prediction, and semantic segmentation, are combined at \emph{each}
stage, thus forming a joint multi-stage processing pipeline. In this way,
the refinement at each stage benefits from the reciprocal relations among
these tasks. Moreover, contextual information is leveraged through an
additional branch for stuff segmentation and a direction path is added to
allow direct information flow across stages. 


\section{Hybrid Task Cascade}
\label{sec:methodology}

\begin{figure*}
	\centering
	\begin{subfigure}[b]{0.45\textwidth}
		\centering
		\includegraphics[width=\textwidth]{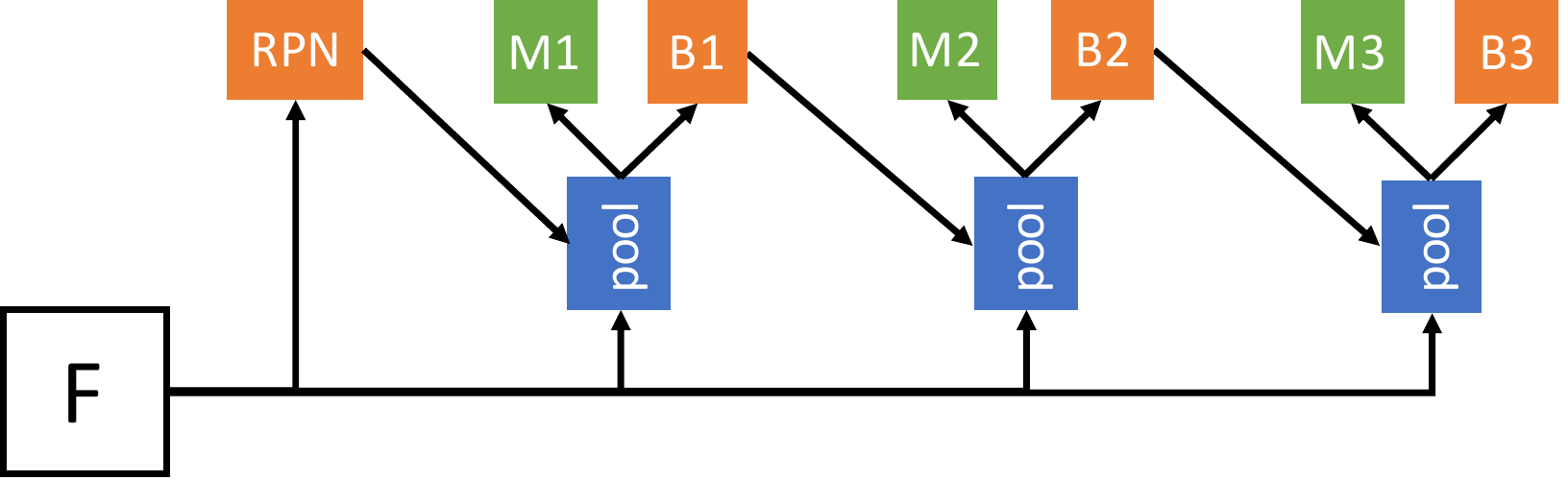}
		\caption{\small Cascade Mask R-CNN}
		\label{fig:framework_a}
	\end{subfigure}
	\hfill
	\begin{subfigure}[b]{0.45\textwidth}
		\centering
		\includegraphics[width=\textwidth]{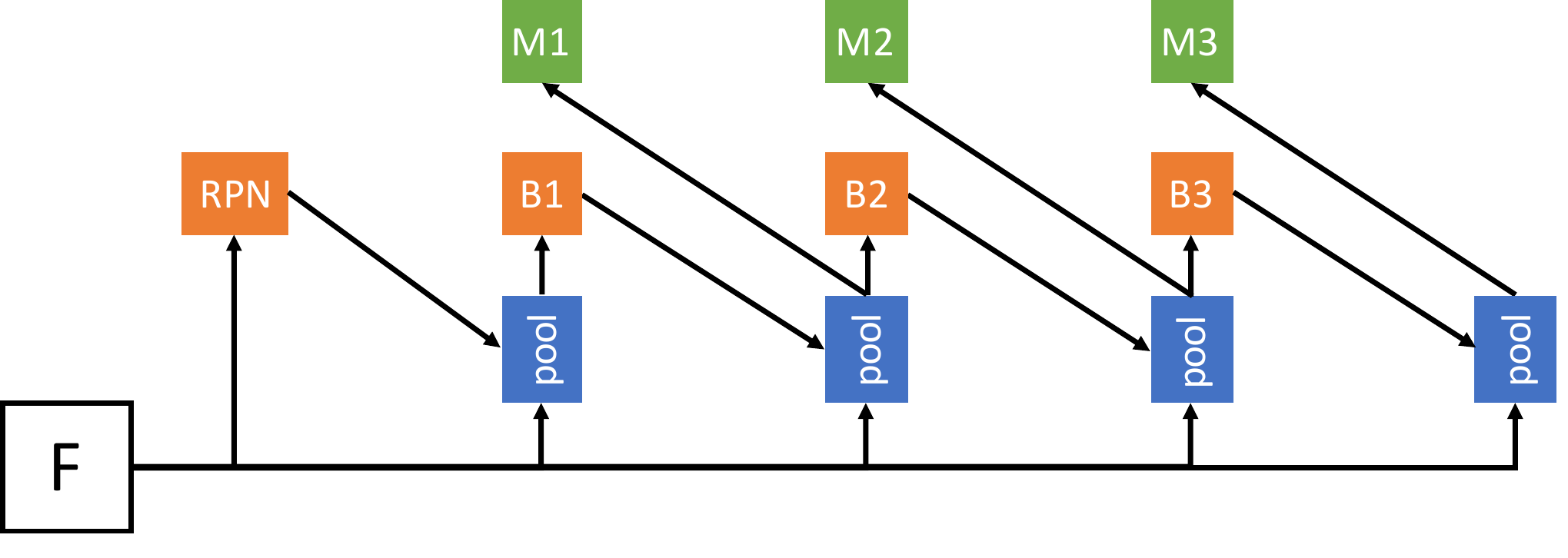}
		\caption{\small Interleaved execution}
		\label{fig:framework_b}
	\end{subfigure}
	\vskip\baselineskip
	\begin{subfigure}[b]{0.45\textwidth}
		\centering
		\includegraphics[width=\textwidth]{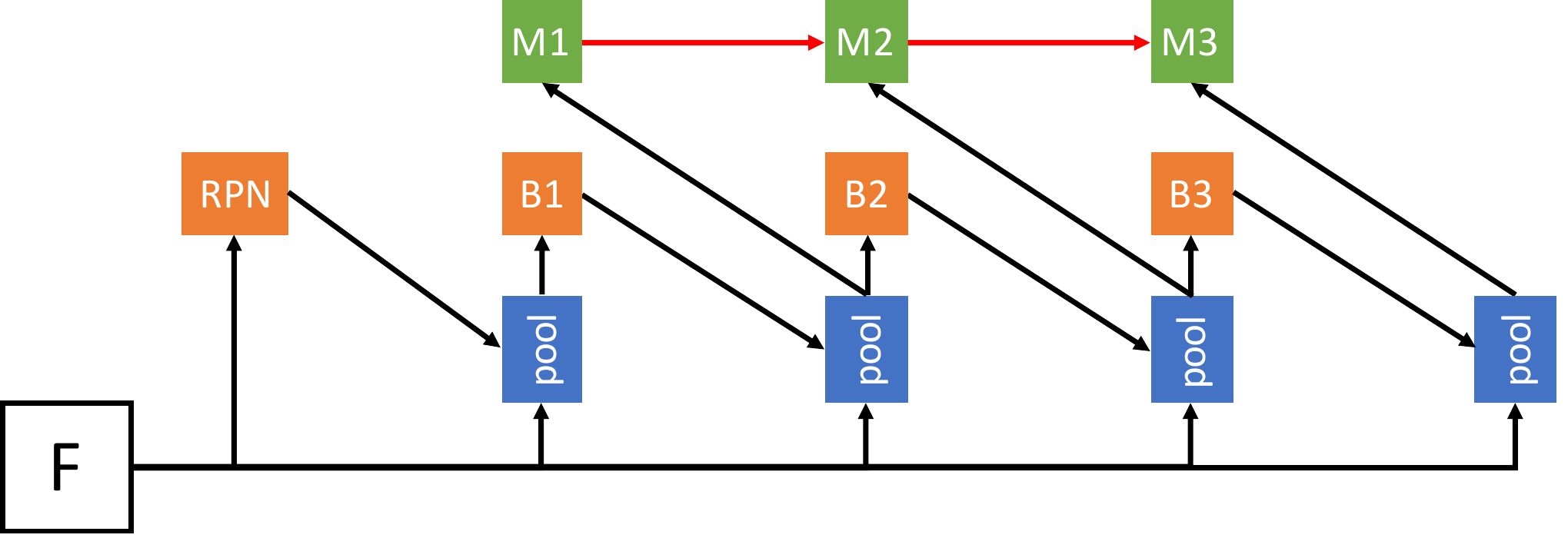}
		\caption{\small Mask information flow}
		\label{fig:framework_c}
	\end{subfigure}
	\hfill
	\begin{subfigure}[b]{0.45\textwidth}
		\centering
		\includegraphics[width=\textwidth]{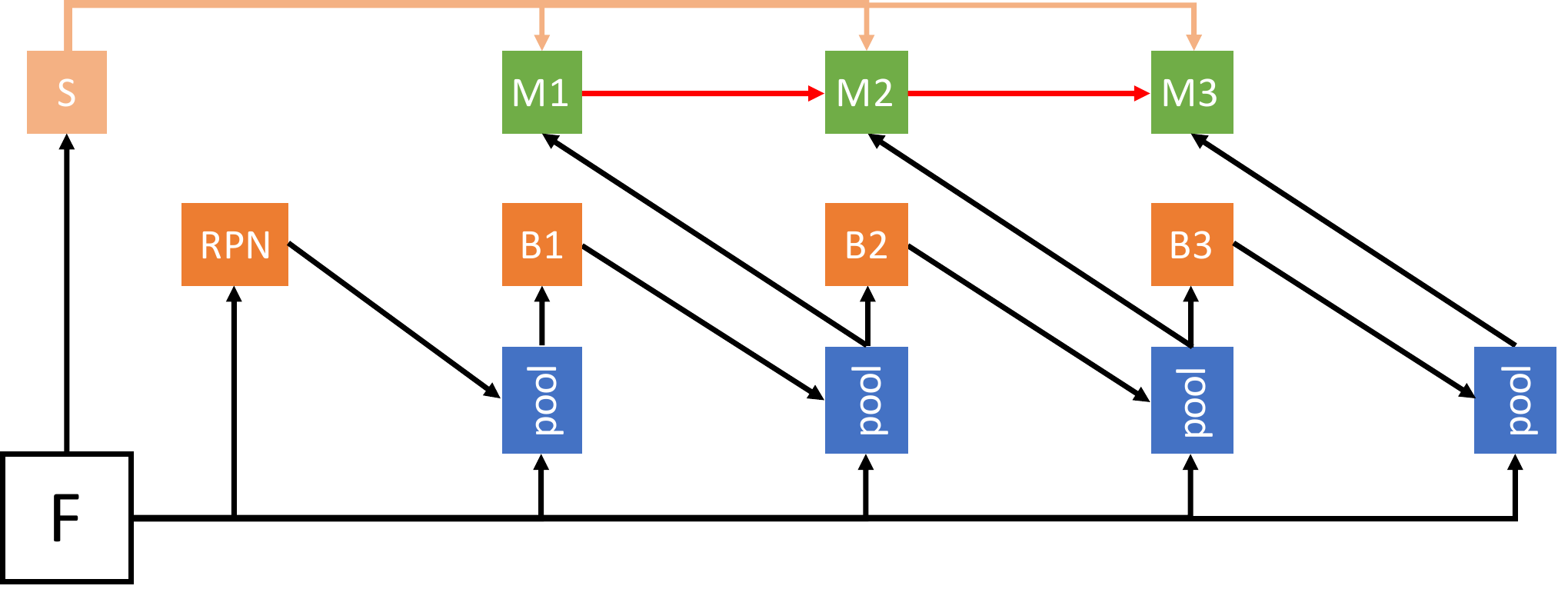}
		\caption{\small Hybrid Task Cascade (semantic feature fusion with box branches is not shown on the figure for neat presentation.)}
		\label{fig:framework_d}
	\end{subfigure}
	\vspace{-0.2cm}
	\caption{The architecture evolution from Cascade Mask R-CNN to Hybrid Task Cascade.}
\end{figure*}

Cascade demonstrated its effectiveness on various tasks such as object
detection~\cite{cai18cascadercnn}. However, it is non-trivial to design a
successful architecture for instance segmentation.
In this work, we find that the key to a successful instance segmentation
cascade is to fully leverage the reciprocal relationship between detection
and segmentation.

\vspace{2pt}
\noindent
\textbf{Overview.}
In this work, we propose Hybrid Task Cascade (HTC), a new framework of
instance segmentation. Compared to existing frameworks, it is distinctive
in several aspects:
(1) It \emph{interleaves} bounding box regression and mask prediction
instead of executing them in parallel.
(2) It incorporates a direct path to reinforce the information flow
between mask branches by feeding the mask features of the preceding
stage to the current one.
(3) It aims to explore more contextual information by adding an additional
semantic segmentation branch and fusing it with box and mask branches.
Overall, these changes to the framework architecture effectively improve
the information flow, not only across stages but also between tasks.

\subsection{Multi-task Cascade}
\label{subsec:multi-task}

\begin{figure}
	\centering
	\includegraphics[width=\linewidth]{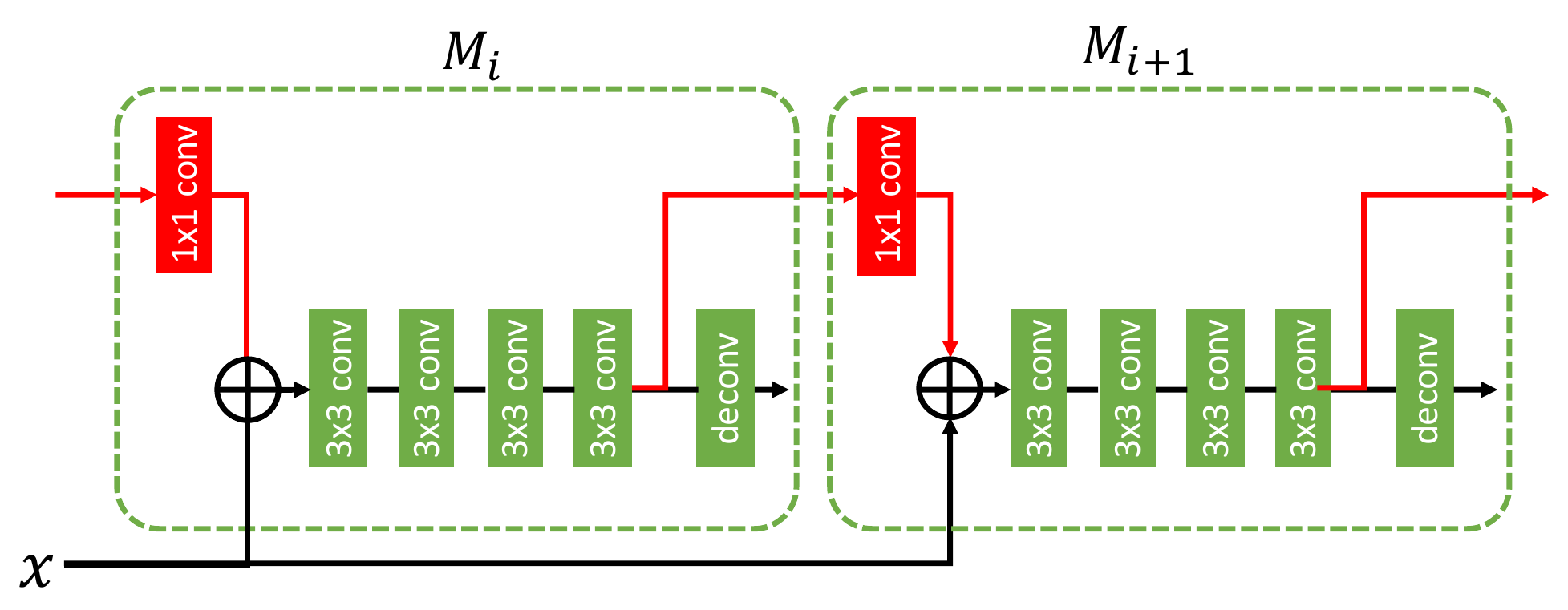}
	\caption{Architecture of multi-stage mask branches.}
	\label{fig:mask-arch}
\end{figure}

\paragraph{Cascade Mask R-CNN.}
We begin with a direct combination of Mask R-CNN and Cascade R-CNN,
denoted as Cascade Mask R-CNN. Specifically, a mask branch following the
architecture of Mask R-CNN is added to each stage of Cascade R-CNN, as shown in
Figure~\ref{fig:framework_a}.
The pipeline is formulated as:
\begin{equation}
	\label{equ:framework_a}
	\begin{split}
		\vx_t^{box} = \cP(\vx, \vr_{t-1})&,  \quad \vr_t = B_t(\vx_t^{box}), \\
		\vx_t^{mask} = \cP(\vx, \vr_{t-1})&,  \quad \vm_t = M_t(\vx_t^{mask}).
	\end{split}
\end{equation}
Here, $\vx$ indicates the CNN features of backbone network,
$\vx_t^{box}$ and $\vx_t^{mask}$ indicates box and mask features derived from
$\vx$ and the input RoIs.
$\cP(\cdot)$ is a pooling operator, \eg, RoI Align or ROI pooling,
$B_t$ and $M_t$ denote the box and mask head at the $t$-th stage,
$\vr_t$ and $\vm_t$ represent the corresponding box predictions and
mask predictions.
By combining the advantages of cascaded refinement and the mutual benefits
between bounding box and mask predictions, this design improves
the box AP, compared to Mask R-CNN and Cascade R-CNN alone.
However, the mask prediction performance remains unsatisfying.

\vspace{-7pt}
\paragraph{Interleaved Execution.}
One drawback of the above design is that the two branches at each stage
are executed \emph{in parallel} during training, both taking the bounding box predictions
from the preceding stage as input. Consequently, the two branches are not
directly interacted within a stage.
In response to this issue, we explore an improved design, which
\emph{interleaves} the box and mask branches, as illustrated in
Figure~\ref{fig:framework_b}.
The interleaved execution is expressed as:
\begin{equation}
	\label{equ:framework_b}
	\begin{split}
		\vx_t^{box} = \cP(\vx, \vr_{t-1})&,  \quad \vr_t = B_t(\vx_t^{box}), \\
		\vx_t^{mask} = \cP(\vx, \vr_t)&,  \quad \vm_t = M_t(\vx_t^{mask}).
	\end{split}
\end{equation}
In this way, the mask branch can take advantage of the updated
bounding box predictions. We found that this yields improved performance.

\vspace{-7pt}
\paragraph{Mask Information Flow.}
In the design above, the mask prediction at each stage is based purely on
the ROI features $\vx$ and the box prediction $\vr_t$. There is no
direct information flow between mask branches at different stages,
which prevents further improvements on mask prediction accuracy.
Towards a good design of mask information flow, we first recall the design of
the cascaded box branches in Cascade R-CNN~\cite{cai18cascadercnn}.
An important point is the input feature of box branch is jointly determined by
the output of the preceding stage and backbone.
Following similar principles, we introduce an information flow between mask
branches by feeding the mask features of the preceding stage to the current stage,
as illustrated in Figure~\ref{fig:framework_c}.
With the direct path between mask branches, the pipeline can be written as:
\begin{equation}
	\label{equ:framework_c}
	\begin{split}
		\vx_t^{box} = \cP(\vx, \vr_{t-1})&, \quad \vr_t = B_t(\vx_t^{box}), \\
		\vx_t^{mask} = \cP(\vx, \vr_t)&, \quad \vm_t = M_t(\cF(\vx_t^{mask},\vm^-_{t-1})),
	\end{split}
\end{equation}
where $\vm^-_{t-1}$ denotes the intermediate feature of $M_{t-1}$ and we use it as
the mask representation of stage $t-1$.
$\cF$ is a function to combine the features of the current stage and the preceding one.
This information flow makes it possible for progressive refinement of masks,
instead of predicting masks on progressively refined bounding boxes.

\vspace{-7pt}
\paragraph{Implementation.}
Following the discussion above, we propose a simple implementation
as below.
\begin{equation}
	\label{equ:bbox-feature}
	\cF(\vx_t^{mask},\vm_{t-1}) = \vx_t^{mask} + \cG_t(\vm^-_{t-1})
\end{equation}
In this implementation, we adopt the RoI feature before the deconvolutional
layer as the mask representation $\vm^-_{t-1}$, whose spatial size is $14\times14$.
At stage $t$, we need to forward all preceding mask heads with RoIs of the current stage to compute $\vm^-_{t-1}$.
\begin{equation}
	\label{equ:recursive-mask}
	\begin{split}
		\vm^-_1 &= M_1^-(\vx_t^{mask}), \\
		\vm^-_2 &= M_2^-(\cF(\vx_t^{mask},\vm^-_1)), \\
		&\vdots\\
		\vm^-_{t-1} &= M_t^-(\cF(\vx_t^{mask},\vm^-_{t-2})).
	\end{split}
\end{equation}
Here, $M_t^-$ denotes the feature transformation component of the
mask head $M_t$, which is comprised of $4$ consecutive $3\times3$ convolutional
layers, as shown in Figure~\ref{fig:mask-arch}.
The transformed features $\vm_{t-1}^-$ are then embedded with
a $1\times 1$ convolutional layer $\cG_t$ in order to be aligned with
the pooled backbone features $\vx_t^{mask}$.
Finally, $\cG_t(\vm_{t-1}^-)$ is added to $\vx_t^{mask}$ through element-wise
sum.
With this introduced bridge, adjacent mask branches are brought into direct
interaction. Mask features in different stages are no longer isolated and
all get supervised through backpropagation.

\subsection{Spatial Contexts from Segmentation}
\label{subsec:contexts}

\begin{figure}
	\centering
	\includegraphics[width=\linewidth]{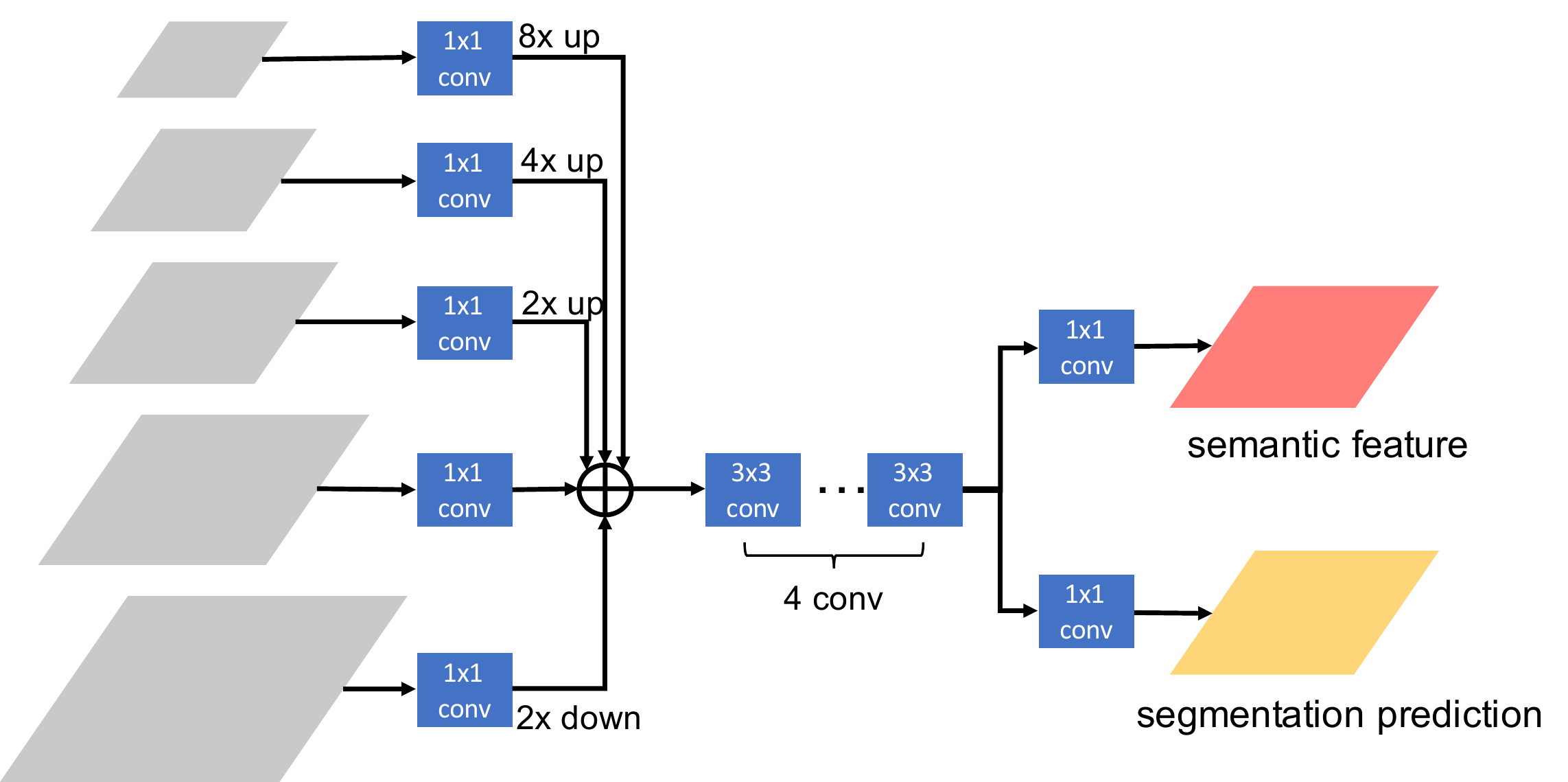}
	\caption{We introduce complementary contextual information by adding semantic segmentation branch.}
	\label{fig:semantic-branch}
\end{figure}

To further help distinguishing the foreground from the cluttered background,
we use the spatial contexts as an effective cue.
We add an additional branch to predict per-pixel semantic segmentation for
the whole image, which adopts the fully convolutional architecture and is
jointly trained with other branches, as shown in Figure~\ref{fig:framework_d}.
The semantic segmentation feature is a strong complement to existing
box and mask features, thus we combine them together for better predictions:
\begin{equation}
	\label{equ:framework_d}
	\begin{split}
		\vx_t^{box} &= \cP(\vx, \vr_{t-1}) + \cP(S(\vx), \vr_{t-1}), \\
		\vr_t &= B_t(\vx_t^{box}), \\
		\vx_t^{mask} &= \cP(\vx, \vr_t) + \cP(S(\vx), \vr_t), \\
		\vm_t &= M_t(\cF(\vx_t^{mask},\vm^-_{t-1})),
	\end{split}
\end{equation}
where $S$ indicates the semantic segmentation head.
In the above formulation, the box and mask heads of each stage take
not only the RoI features extracted from the backbone as input, but also
exploit semantic features, which can be more discriminative on cluttered
background.

\vspace{-7pt}
\paragraph{Semantic Segmentation Branch.}
Specifically, the semantic segmentation branch $S$ is constructed
based on the output of the Feature Pyramid~\cite{lin2017feature}.
Note that for semantic segmentation, the features at a single level
may not be able to provide enough discriminative power. Hence,
our design incorporates the features at multiple levels.
In addition to the mid-level features, we also incorporate higher-level
features with global information and lower-level features with local information
for better feature representation.


Figure~\ref{fig:semantic-branch} shows the architecture of this branch.
Each level of the feature pyramid is first aligned to a common representation
space via a $1\times 1$ convolutional layer.
Then low level feature maps are upsampled, and high level feature maps are
downsampled to the same spatial scale, where the stride is set to $8$.
We found empirically that this setting is sufficient for fine pixel-level
predictions on the whole image.
These transformed feature maps from different levels are subsequently fused
by element-wise sum.
Moreover, we add four convolutional layers thereon to further bridge the
semantic gap.
At the end, we simply adopt a convolutional layer to predict the pixel-wise
segmentation map.
Overall, we try to keep the design of semantic segmentation branch simple
and straightforward.
Though a more delicate structure can further improve the performance,
It goes beyond our scope and we leave it for future work.

\vspace{0.1cm}
\noindent
\textbf{Fusing Contexts Feature into Main Framework.}
It is well known that joint training of closely related tasks can improve
feature representation and bring performance gains to original tasks.
Here, we propose to fuse the semantic features with box/mask features to allow
more interaction between different branches. In this way, the semantic branch
\emph{directly} contributes to the prediction of bounding boxes and masks
with the encoded spatial contexts.
Following the standard practice, given a RoI, we use RoIAlign to extract
a small (\eg, $7\times7$ or $14\times 14$) feature patch from the corresponding
level of feature pyramid outputs as the representation. At the same time, we
also apply RoIAlign on the feature map of the semantic branch and obtain
a feature patch of the same shape, and then combine the features from both
branches by element-wise sum.

\subsection{Learning}
\label{subsec:training}

Since all the modules described above are differentiable, Hybrid Task Cascade
(HTC) can be trained in an end-to-end manner. At each stage $t$, the box head
predicts the classification score $c_t$ and regression offset $\vr_t$ for all
sampled RoIs. The mask head predicts pixel-wise masks $\vm_t$ for positive RoIs.
The semantic branch predicts a full image semantic segmentation map $\mathbf{s}$.
The overall loss function takes the form of a multi-task learning:
\begin{equation}
	\label{equ:loss}
	\begin{split}
		&L = \sum_{t=1}^T \alpha_t (L_{bbox}^t + L_{mask}^t) + \beta L_{seg}, \\
		&L_{bbox}^t(c_i, \vr_t, \hat{c}_t, \hat{\vr}_t) = L_{cls}(c_t, \hat{c}_t) + L_{reg}(\vr_t, \hat{\vr}_t), \\
		&L_{mask}^t(\vm_t, \hat{\vm}_t) = \text{BCE}(\vm_t, \hat{\vm}_t), \\
		&L_{seg} = CE(\mathbf{s}, \hat{\mathbf{s}}).
	\end{split}
\end{equation}
Here,
$L_{bbox}^t$ is the loss of the bounding box predictions at stage $t$, which
follows the same definition as in Cascade R-CNN~\cite{cai18cascadercnn} and combines two
terms $L_{cls}$ and $L_{reg}$, respectively for classification and bounding
box regression.
$L_{mask}^t$ is the loss of mask prediction at stage $t$,
which adopts the binary cross entropy form as in Mask R-CNN~\cite{he2017mask}.
$L_{seg}$ is the semantic segmentation loss in the form of cross entropy.
The coefficients $\alpha_t$ and $\beta$ are used to balance the contributions
of different stages and tasks.
We follow the hyperparameter settings in Cascade R-CNN~\cite{cai18cascadercnn}.
Unless otherwise noted, we set $\alpha = [1, 0.5, 0.25]$, $T=3$ and
$\beta = 1$ by default.


\section{Experiments}
\label{sec:experiments}

\begin{table*}[htb]
	\centering
	\caption{Comparison with state-of-the-art methods on COCO test-dev dataset.}
	\vspace{0.1cm}
	\addtolength{\tabcolsep}{-1pt}
	\begin{tabular}{*{10}{c}}
		\toprule
		Method                       & Backbone        & box AP & mask AP & $\text{AP}_{50}$ & $\text{AP}_{75}$ & $\text{AP}_{S}$ & $\text{AP}_{M}$ & $\text{AP}_{L}$ & runtime (fps) \\
		\midrule
		Mask R-CNN~\cite{he2017mask} & ResNet-50-FPN   & 39.1   & 35.6    & 57.6             & 38.1             & 18.7            & 38.3            & 46.6            & 5.3           \\
		PANet\cite{liu2018path}      & ResNet-50-FPN   & 41.2   & 36.6    & 58.0             & 39.3             & 16.3            & 38.1            & 52.4            & -             \\
		\midrule
		Cascade Mask R-CNN           & ResNet-50-FPN   & 42.7   & 36.9    & 58.6             & 39.7             & 19.6            & 39.3            & 48.8            & 3.0           \\
		Cascade Mask R-CNN           & ResNet-101-FPN  & 44.4   & 38.4    & 60.2             & 41.4             & 20.2            & 41.0            & 50.6            & 2.9           \\
		Cascade Mask R-CNN           & ResNeXt-101-FPN & 46.6   & 40.1    & 62.7             & 43.4             & 22.0            & 42.8            & 52.9            & 2.5           \\
		\midrule
		HTC (ours)                   & ResNet-50-FPN   & \textbf{43.6}   & \textbf{38.4}    & 60.0             & 41.5             & 20.4            & 40.7            & 51.2            & 2.5           \\
		HTC (ours)                   & ResNet-101-FPN  & \textbf{45.3}   & \textbf{39.7}    & 61.8             & 43.1             & 21.0            & 42.2            & 53.5            & 2.4           \\
		HTC (ours)                   & ResNeXt-101-FPN & \textbf{47.1}   & \textbf{41.2}    & 63.9             & 44.7             & 22.8            & 43.9            & 54.6            & 2.1           \\
		\bottomrule
	\end{tabular}
	\label{tab:overall-results}
\end{table*}

\subsection{Datasets and Evaluation Metrics}

\noindent
\textbf{Datasets.}
We perform experiments on the challenging COCO dataset~\cite{lin2014microsoft}.
We train our models on the split of 2017\emph{train} (115k images)
and report results on 2017\emph{val} and 2017\emph{test-dev}.
Typical instance annotations are used to supervise box and mask branches,
and the semantic branch is supervised by COCO-stuff~\cite{caesar2018cvpr} annotations.

\noindent
\textbf{Evaluation Metrics.}
We report the standard COCO-style Average Precision (AP) metric which averages APs across IoU thresholds from $0.5$ to $0.95$ with an interval of $0.05$.
Both box AP and mask AP are evaluated.
For mask AP, we also report $\text{AP}_{50}$, $\text{AP}_{75}$ (AP at different IoU thresholds)
and $\text{AP}_{S}$, $\text{AP}_{M}$, $\text{AP}_{L}$ (AP at different scales).
Runtime is measured on a single TITAN Xp GPU.

\subsection{Implementation Details}
In all experiments, we adopt a 3-stage cascade. FPN is used in all backbones.
For fair comparison, Mask R-CNN and Cascade R-CNN are reimplemented with PyTorch~\cite{paszke2017automatic} and mmdetection~\cite{mmdetection2018},
which are slightly higher than the reported performance in the original papers.
We train detectors with 16 GPUs (one image per GPU) for $20$ epoches with an initial learning rate of $0.02$,
and decrease it by $0.1$ after $16$ and $19$ epoches, respectively.
The long edge and short edge of images are resized to $1333$ and $800$ respectively without changing the aspect ratio.

During inference, object proposals are refined progressively by box heads of different stages.
Classification scores of multiple stages are ensembled as in Cascade R-CNN.
Mask branches are only applied to detection boxes with higher scores than a threshold (0.001 by default).

\subsection{Benchmarking Results}

We compare HTC with the state-of-the-art instance segmentation approaches on the COCO dataset in Table~\ref{tab:overall-results}.
We also evaluate Cascade Mask R-CNN, which is described in Section~\ref{sec:intro}, as a strong baseline of our method.
Compared to Mask R-CNN, the naive cascaded baseline brings $3.5\%$ and $1.2\%$ gains in terms of box AP and mask AP respectively.
It is noted that this baseline is already higher than PANet~\cite{liu2018path}, the state-of-the-art instance segmentation method.
Our HTC achieves consistent improvements on different backbones, proving its effectiveness.
It achieves a gain of $1.5\%$, $1.3\%$ and $1.1\%$ for ResNet-50, ResNet-101 and ResNeXt-101, respectively.


\subsection{Ablation Study}

\noindent\textbf{Component-wise Analysis.}
Firstly, we investigate the effects of main components in our framework.
``Interleaved'' denotes the interleaved execution of bbox and mask branches,
``Mask Info'' indicates the mask branch information flow and
``Semantic'' means introducing the semantic segmentation branch.
From Table~\ref{tab:components}, we can learn that the interleaved execution
slightly improves the mask AP by $0.2\%$.
The mask information flow contributes to a further $0.6\%$ improvement,
and the semantic segmentation branch leads to a gain of $0.6\%$.

\begin{table*}[htb]
	\centering
	\caption{Effects of each component in our design. Results are reported on COCO 2017 val.}
	\begin{tabular}{*{4}{c}|*{7}{c}}
		\toprule
		Cascade    & Interleaved & Mask Info  & Semantic   & box AP & mask AP & $\text{AP}_{50}$ & $\text{AP}_{75}$ & $\text{AP}_{S}$ & $\text{AP}_{M}$ & $\text{AP}_{L}$ \\
		\hline
		\checkmark &             &            &            & 42.5   & 36.5    & 57.9             & 39.4             & 18.9            & 39.5            & 50.8           \\
		\checkmark & \checkmark  &            &            & 42.5   & 36.7    & 57.7             & 39.4             & 18.9            & 39.7            & 50.8           \\
		\checkmark & \checkmark  & \checkmark &            & 42.5   & 37.4    & 58.1             & 40.3             & 19.6            & 40.3            & 51.5           \\
		\checkmark & \checkmark  & \checkmark & \checkmark & 43.2   & \textbf{38.0}    & 59.4             & 40.7             & 20.3            & 40.9            & 52.3           \\
		\bottomrule
	\end{tabular}
	\label{tab:components}
\end{table*}

\noindent\textbf{Effectiveness of Interleaved Branch Execution.}
In Section~\ref{subsec:multi-task}, we design the interleaved branch execution
to benefit the mask branch from updated bounding boxes during training.
To investigate the effeciveness of this strategy, we compare it with the
conventional parallel execution pipeline on both Mask R-CNN and Cascade Mask R-CNN.
As shown in Table~\ref{tab:interleaved}, interleaved execution outperforms 
parallel execution on both methods, with an improvement of $0.5\%$ and $0.2\%$
respectively.

\begin{table*}[htb]
	\centering
	\caption{Results of parallel/interleaved branch execution on different methods.}
	\begin{tabular}{*{9}{c}}
		\toprule
		Method                              & execution   & box AP & mask AP & $\text{AP}_{50}$ & $\text{AP}_{75}$ & $\text{AP}_{S}$ & $\text{AP}_{M}$ & $\text{AP}_{L}$ \\
		\midrule
		\multirow{2}{*}{Mask R-CNN}         & parallel    &   38.4   &  35.1  &  56.6   &       37.4       &     18.7         &    38.4         &       47.7                      \\
		                                    & interleaved &   38.7 &   \textbf{35.6}  &     57.2         &       37.9        &    19.0        &      39.0       &      48.3       \\
		\midrule
		\multirow{2}{*}{Cascade Mask R-CNN} & parallel    &  42.5  &  36.5   &       57.9       &    39.4          &     18.9        &     39.5       &     50.8    \\
		                                    & interleaved &  42.5  &  \textbf{36.7}   &       57.7       &    39.4          &      18.9      &      39.7       &       50.8      \\
		\bottomrule
	\end{tabular}
	\label{tab:interleaved}
\end{table*}

\noindent\textbf{Effectiveness of Mask Information Flow.}
We study how the introduced mask information flow helps mask prediction by
comparing stage-wise performance.
Semantic segmentation branch is not involved to exclude possible distraction.
From Table~\ref{tab:mask-info-stagewise}, we find that introducing the mask
information flow greatly improves the the mask AP in the second stage.
Without direct connections between mask branches, the second stage only benefits
from better localized bounding boxes, so the improvement is limited ($0.8\%$).
With the mask information flow, the gain is more significant ($1.5\%$),
because it makes each stage aware of the preceding stage's features.
Similar to Cascade R-CNN, stage 3 does not outperform stage 2, but it
contributes to the ensembled results.

\begin{table}[htb]
	\centering
	\caption{Effects of the mask information flow. We evaluate the stage-wise and ensembled performance with or without the information flow (denoted as I.F.).}
	\addtolength{\tabcolsep}{-2pt}
	\begin{tabular}{*{8}{c}}
		\toprule
		I.F. & test stage & AP & $\text{AP}_{50}$ & $\text{AP}_{75}$ & $\text{AP}_{S}$ & $\text{AP}_{M}$ & $\text{AP}_{L}$ \\
		\midrule
		\multirow{4}{*}{N} & stage 1    & 35.5 & 56.7 & 37.8 & 18.7 & 38.8 & 48.6 \\
		       & stage 2    & 36.3 & 57.5 & 39.0 & 18.8 & 39.4 & 50.6 \\
		       & stage 3    & 35.9 & 56.5 & 38.7 & 18.2 & 39.1 & 49.9 \\
		       & stage $\overline{1\sim3}$ & 36.7 & 57.7 & 39.4 & 18.9 & 39.7 & 50.8 \\
		\midrule
		\multirow{4}{*}{Y} & stage 1    & 35.5 & 56.8 & 37.8 & 19.0 & 38.8 & 49.0 \\
		       & stage 2    & 37.0 & 58.0 & 39.8 & 19.4 & 39.8 & 51.3 \\
		       & stage 3    & 36.8 & 57.2 & 39.9 & 18.7 & 39.8 & 51.1 \\
		       & stage $\overline{1\sim3}$ & \textbf{37.4} & 58.1 & 40.3 & 19.6 & 40.3 & 51.5 \\
		\bottomrule
	\end{tabular}
	\label{tab:mask-info-stagewise}
\end{table}

\noindent\textbf{Effectiveness of Semantic Feature Fusion.}
We exploit contextual features by introducing a semantic segmentation branch and fuse the features of different branches.
Multi-task learning is known to be beneficial, here we study the necessity of semantic feature fusion.
We train different models that fuse semantic features with the box or mask or both branches,
and the results are shown in Table~\ref{tab:semantic}.
Simply adding a full image segmentation task achieves $0.6\%$ improvement, mainly resulting from additional supervision.
Feature fusion also contributes to further gains,\eg, fusing the semantic features
with both the box and mask branches brings an extra $0.4\%$ gain, which indicates that complementary information increases feature discrimination for box and mask branches.

\begin{table}[htb]
	\centering
	\caption{Ablation study of semantic feature fusion on COCO 2017 val.}
	\addtolength{\tabcolsep}{-2pt}
	\begin{tabular}{*{7}{c}}
		\toprule
		Fusion & AP   & $\text{AP}_{50}$ & $\text{AP}_{75}$ & $\text{AP}_{S}$ & $\text{AP}_{M}$ & $\text{AP}_{L}$ \\
		\midrule
		-             & 36.5 & 57.9 & 39.4 & 18.9 & 39.5 & 50.8 \\
		none          & 37.1 & 58.6 & 39.9 & 19.3 & 40.0 & 51.7 \\
		bbox          & 37.3 & 58.9 & 40.2 & 19.4 & 40.2 & 52.3 \\
		mask          & 37.4 & 58.7 & 40.2 & 19.4 & 40.1 & 52.4 \\
		both          & \textbf{37.5} & 59.1 & 40.4 & 19.6 & 40.3 & 52.6 \\
		\bottomrule
	\end{tabular}
	\vspace{-0.3cm}
	\label{tab:semantic}
\end{table}

\noindent\textbf{Influence of Loss Weight.}
The new hyper-parameter $\beta$ is introduced, since we involve one more task for joint training.
We tested different loss weight for the semantic branch, as shown in Table~\ref{tab:loss-weight}.
Results show that our method is insensitive to the loss weight.

\begin{table}[htb]
	\centering
	\caption{Ablation study of semantic branch loss weight $\beta$ on COCO 2017 val.}
	\addtolength{\tabcolsep}{-2pt}
	\begin{tabular}{*{7}{c}}
		\toprule
		$\beta$ & AP & $\text{AP}_{50}$ & $\text{AP}_{75}$ & $\text{AP}_{S}$ & $\text{AP}_{M}$ & $\text{AP}_{L}$ \\
		\midrule
		0.5     &37.9&       59.3       &      40.7        &      19.7       &      41.0       &       52.5      \\
		1       &\textbf{38.0}&       59.4       &      40.7        &      20.3       &      40.9       &       52.3      \\
		2       &37.9&       59.3       &      40.6        &      19.6       &      40.8       &       52.8      \\
		3       &37.8&       59.0       &      40.5        &      19.9       &      40.5       &       53.2      \\
		\bottomrule
	\end{tabular}
	\label{tab:loss-weight}
\end{table}

\subsection{Extensions on HTC}

With the proposed HTC, we achieve $49.0$ mask AP and $2.3\%$ absolute improvement compared to the winning entry last year.
Here we list all the steps and additional modules used to obtain the performance.
The step-by-step gains brought by each component are illustrated in Table~\ref{tab:challenge-results}.

\noindent\textbf{HTC Baseline.}
The ResNet-50 baseline achieves $38.2$ mask AP.

\noindent\textbf{DCN.}
We adopt deformable convolution~\cite{dai2017deformable} in the last stage (res5) of the backbone.

\noindent\textbf{SyncBN.}
Synchronized Batch Normalization~\cite{peng2018megdet,liu2018path} is used in the backbone and heads.

\noindent\textbf{Multi-scale Training.}
We adopt multi-scale training. In each iteration, the scale of short edge is randomly sampled from $[400, 1400]$, and the scale of long edge is fixed as $1600$.

\noindent\textbf{SENet-154.}
We tried different backbones besides ResNet-50, and SENet-154~\cite{Hu_2018_CVPR} achieves best single model performance among them.

\noindent\textbf{GA-RPN.}
We finetune trained detectors with the proposals generated by GA-RPN~\cite{wang2019region},
which achieves near 10\% higher recall than RPN.

\noindent\textbf{Multi-scale Testing.}
We use 5 scales as well as horizontal flip at test time and ensemble the results.
Testing scales are (600, 900), (800, 1200), (1000, 1500), (1200, 1800), (1400, 2100).

\noindent\textbf{Ensemble.}
We utilize an emsemble of five networks: SENet-154~\cite{Hu_2018_CVPR}, ResNeXt-101~\cite{xie2017aggregated} 64x4d, ResNeXt-101 32x8d, DPN-107~\cite{chen2017dual}, FishNet~\cite{fishnet_18_nips}.

\begin{table}[htb]
	\centering
	\caption{Results (mask AP) with better backbones and bells and whistles on COCO test-dev dataset.}
	\addtolength{\tabcolsep}{-2pt}
	\begin{tabular}{*{9}{c}}
		\toprule
		                              & AP            & $\text{AP}_{50}$ & $\text{AP}_{75}$ & $\text{AP}_{S}$ & $\text{AP}_{M}$ & $\text{AP}_{L}$ \\
		\midrule
		2017 winner \cite{liu2018path} & 46.7          & 69.5             & 51.3             & 26.0            & 49.1            & \textbf{64.0}   \\
		Ours                          & \textbf{49.0} & \textbf{73.0}    & \textbf{53.9}    & \textbf{33.9}   & \textbf{52.3}   & 61.2            \\
		\midrule
		HTC baseline                  & 38.4          & 60.0             & 41.5             & 20.4            & 40.7            & 51.2            \\
		+ DCN                         & 39.5          & 61.3             & 42.8             & 20.9            & 41.8            & 52.7            \\
		+ SyncBN                      & 40.7          & 62.8             & 44.2             & 22.2            & 43.1            & 54.4            \\
		+ ms train                    & 42.5          & 64.8             & 46.4             & 23.7            & 45.3            & 56.7            \\
		+ SENet-154                   & 44.3          & 67.5             & 48.3             & 25.0            & 47.5            & 58.9            \\
		+ GA-RPN                      & 45.3          & 68.9             & 49.4             & 27.0            & 48.3            & 59.6            \\
		+ ms test                     & 47.4          & 70.6             & 52.1             & 30.2            & 50.1            & 61.8            \\
		+ ensemble                    & 49.0          & 73.0             & 53.9             & 33.9            & 52.3            & 61.2            \\
		\bottomrule
	\end{tabular}
	\vspace{-0.3cm}
	\label{tab:challenge-results}
\end{table}

\begin{figure*}
	\centering
	\includegraphics[width=\linewidth]{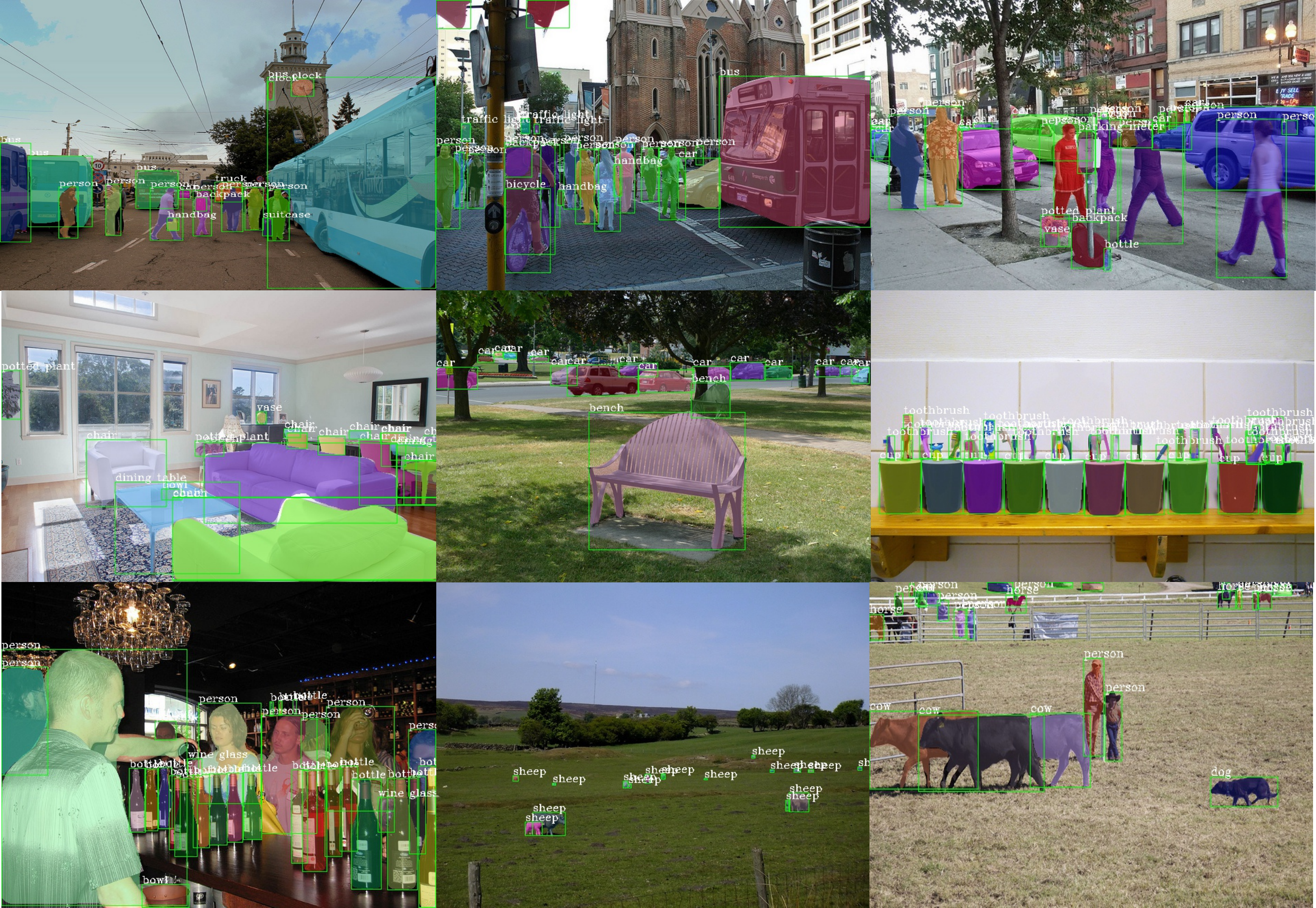}
	\caption{Examples of segmentation results on COCO dataset.}
	\vspace{-0.3cm}
	\label{fig:examples}
\end{figure*}

\subsection{Extensive Study on Common Modules}
We also perform extensive study on some components designed for detection and segmentation.
Components are often compared under different conditions such as backbones, codebase, \etc.
Here we provide a unified environment with state-of-the-art object detection and
instance segmentation framework to investigate the functionality of extensive components.
We integrate several common modules designed for detection and segmentation
and evaluate them under the same settings, and the results are shown in Table~\ref{tab:extensive-study}.
Limited by our experience and resources, some implementations and the integration
methods may not be optimal and worth further study.
Code will be released as a benchmark to test more components.

\noindent\textbf{ASPP.}
We adopt the Atrous Spatial Pyramid Pooling (ASPP)~\cite{chen2018deeplab} module
from the semantic segmentation community to capture more image context at multiple scales.
We append an ASPP module after FPN.

\noindent\textbf{PAFPN.}
We test the PAFPN module from PANet~\cite{liu2018path}. The difference from
the original implementation is that we do not use Synchronized BatchNorm.

\noindent\textbf{GCN.}
We adopt Global Convolutional Network (GCN)~\cite{peng2017large} in the semantic segmentation branch.

\noindent\textbf{PreciseRoIPooling.}
We replace the RoI align layers in HTC with Precise RoI Pooling~\cite{jiang2018acquisition}.

\noindent\textbf{SoftNMS.}
We apply SoftNMS~\cite{bodla2017soft} to box results.


\begin{table}[htb]
	\centering
	\caption{Extensive study on related modules on COCO 2017 val.}
	\label{tab:extensive-study}
	\addtolength{\tabcolsep}{-2pt}
	\begin{tabular}{*{7}{c}}
		\toprule
		Method       & AP & $\text{AP}_{50}$ & $\text{AP}_{75}$ & $\text{AP}_{S}$ & $\text{AP}_{M}$ & $\text{AP}_{L}$ \\
		\midrule
		HTC          &38.0&       59.4       &       40.7       &      20.3       &      40.9       &      52.3       \\
		HTC+ASPP     &38.1&       59.9       &       41.0       &      20.0       &      41.2       &      52.8       \\
		HTC+PAFPN    &38.1&       59.5       &       41.0       &      20.0       &      41.2       &      53.0       \\
		HTC+GCN      &37.9&       59.2       &       40.7       &      20.0       &      40.6       &      52.3       \\
		HTC+PrRoIPool&37.9&       59.1       &       40.9       &      19.7       &      40.9       &      52.7       \\
		HTC+SoftNMS  &38.3&       59.6       &       41.2       &      20.4       &      41.2       &      52.7       \\
		\bottomrule
	\end{tabular}
	\vspace{-0.4cm}
\end{table}
\section{Conclusion}

We propose Hybrid Task Cascade (HTC), a new cascade architecture for instance segmentation.
It interweaves box and mask branches for a joint multi-stage processing, and adopts a semantic segmentation branch to provide spatial context.
This framework progressively refines mask predictions and integrates complementary features together in each stage.
Without bells and whistles, the proposed method obtains
$1.5\%$ improvement over a strong Cascade Mask R-CNN baseline on MSCOCO dataset.
Notably, our overall system achieves $48.6$ mask AP on the
test-challenge dataset and $49.0$ mask AP on test-dev.

\vspace{-12pt}
\paragraph{Acknowledgments.}
This work is partially supported by the Collaborative Research grant from SenseTime Group (CUHK Agreement No. TS1610626 \& No. TS1712093), the General Research Fund (GRF) of Hong Kong (No. 14236516, No. 14203518 \& No. 14224316), and Singapore MOE AcRF Tier 1 (M4012082.020).

{\small
\bibliographystyle{ieee_fullname}
\bibliography{sections/egbib}
}

\end{document}